\documentclass[11pt]{article}

\usepackage[letterpaper,margin=1in]{geometry}

\usepackage[T1]{fontenc}
\usepackage[utf8]{inputenc}
\usepackage{mathptmx}          
\usepackage{microtype}

\usepackage{amsmath,amssymb,amsthm,mathtools}
\newtheorem{proposition}{Proposition}
\newcommand{\indfn}{\mathbf{1}}          
\DeclareMathOperator{\KL}{D_{\mathrm{KL}}}
\DeclareMathOperator{\Jac}{Jaccard}
\DeclareMathOperator*{\argmax}{arg\,max}

\usepackage{booktabs}
\usepackage{array}
\usepackage{graphicx}
\usepackage{caption}
\usepackage{xcolor}
\definecolor{r0col}{RGB}{198,224,180}     
\definecolor{r1col}{RGB}{189,215,238}     
\definecolor{boxgray}{RGB}{248,248,248}
\definecolor{linkcol}{RGB}{20,60,120}

\usepackage{tikz}
\usetikzlibrary{positioning,arrows.meta,shapes.geometric,fit,backgrounds,calc}

\usepackage{enumitem}

\usepackage[numbers,sort&compress]{natbib}
\usepackage[colorlinks=true,linkcolor=linkcol,citecolor=linkcol,
            urlcolor=linkcol,breaklinks=true]{hyperref}
\usepackage{url}

\usepackage{titlesec}
\titlespacing*{\section}{0pt}{1.4ex plus 1ex minus .2ex}{1ex}
\titlespacing*{\subsection}{0pt}{1.1ex plus 1ex minus .2ex}{0.7ex}

\title{\bfseries ProvenAI: \\ Provenance-Native Traces of Evidence in Generated Answers}

\author{%
  Mohammad Faizan \qquad Dalal Alharthi\\
  College of Information Science\\
  University of Arizona\\
  \texttt{\{mohammadfaizan, dalharthi\}@arizona.edu}%
}
\date{}

\begin{document}
\maketitle
\thispagestyle{plain}
\begin{center}\small Preprint.\end{center}

\begin{abstract}
Retrieval-augmented systems routinely present citations alongside
generated answers, yet a citation does not confirm that the
corresponding source meaningfully shaped the output. This paper
introduces ProvenAI, a framework that decomposes transparency in
multi-hop question answering into three independently measurable
layers: answer correctness, citation fidelity against benchmark
supporting evidence, and per-document influence under
leave-one-resource-out intervention. Targeting the HotpotQA distractor
benchmark through a seven-stage pipeline covering data normalisation,
retrieval indexing, citation-aware answer generation, attribution
auditing, ablation-based influence estimation, batch evaluation, and
interactive inspection, ProvenAI evaluates 7{,}405 validation examples
drawn from a canonical corpus of 509{,}300 passages. The system
achieves 53.53\% answer accuracy alongside a mean citation-fidelity
score of 71.55\%, and a worked example surfaces what we call the
\emph{citation-influence gap}: a clean citation audit co-occurring with
a profile in which one cited source registers only weak influence while
seven uncited sources demonstrably shift the output. We formalise the
relationship between the implemented surface proxy and a token-level
KL-divergence target through a stated faithfulness condition, ground the
framework in causal-mediation analysis and database-provenance theory,
and discuss how the three measurement layers compose with cryptographic
provenance architectures emerging in autonomous scientific discovery.
ProvenAI establishes that meaningful transparency in retrieval-grounded
QA requires traceable links across retrieved, cited, and behaviourally
influential evidence as three distinct, independently measured layers.
\end{abstract}

\section{Introduction}
Retrieval-augmented generation has become a mainstream mechanism for
improving the factual grounding of language model outputs by supplying
external documents at inference time \citep{lewis2020rag,gao2023ragsurvey}.
Multi-hop question answering amplifies the challenge: a correct answer
typically requires chaining evidence across several documents, making it
essential to understand not only what was retrieved but how each piece of
evidence influenced the final response \citep{yang2018hotpotqa,tang2024multihoprag}.
The HotpotQA distractor benchmark provides sentence-level supporting-fact
annotations and has therefore become a principled testbed for studying
how retrieved evidence propagates through a pipeline.

Despite the promise of citation-bearing outputs, a model that receives
ten retrieved passages and cites only two of them may still be shaped by
the remaining eight. The citation list answers a single narrow question,
namely what the model claims to have used, and says nothing about whether
those sources matched the gold supporting facts, nor whether removing any
particular document would have changed the response. State-of-the-art
systems leave citations unsupported a substantial fraction of the time
and frequently cite sources in ways that do not reflect their actual
context usage \citep{liu2023verifiability,gao2023enabling,qi2024mirage}.
Recent work further shows that language models can be distracted by
irrelevant retrieved context \citep{shi2023distraction} and may rely on
parametric memory rather than the cited passages
\citep{mallen2023whennot}, making the gap between cited and influential
evidence operationally important. The need for output-level auditing is
especially acute in security and forensic applications, where
LLM-driven pipelines increasingly analyse logs and evidence yet must
remain accountable to a verifiable record of what shaped each conclusion
\citep{alharthi2025llmforensics,alharthi2025ciaf}.

ProvenAI is built around this gap. It is not a stronger QA model but an
infrastructure for measuring how evidence propagates through a
retrieval-grounded pipeline. The core premise is that answer quality,
citation faithfulness, and resource influence each capture a different
dimension of transparency, and collapsing them into a single score
discards the diagnostic information practitioners need. The implemented
system follows a seven-stage workflow in which HotpotQA data are
converted into stable local artefacts, a dense retrieval index is built,
a generation step produces citation-aware answers, an auditing step
checks citations against gold supporting facts, leave-one-resource-out
ablations estimate per-document influence, metrics are aggregated over
the full validation split, and a Streamlit dashboard exposes results
alongside Model Context Protocol \citep{anthropic2024mcp} trace
artefacts.

The contributions of this work are fourfold. We present a practical
decomposition of retrieval-grounded QA transparency into three
independently measurable layers and supply formal operationalisations
for each. We report an end-to-end pipeline evaluated over the complete
HotpotQA distractor validation split, with reproducible structured
report artefacts at every stage. We exhibit and characterise the
citation-influence gap, formalise the relationship between the
implemented surface proxy and a token-level KL-divergence target, and
connect the framework to causal-mediation analysis and
database-provenance theory. Finally, we discuss how the three
measurement layers compose with cryptographic provenance architectures
relevant to autonomous scientific discovery, settings in which the audit
trail must be an invariant of execution rather than retrospective
annotation.

\section{Related Work}
\paragraph{Multi-hop retrieval and RAG foundations.}
Retrieval-augmented generation pairs a parametric language model with a
non-parametric retrieval index over a corpus \citep{lewis2020rag}, with
dense passage retrieval \citep{karpukhin2020dpr} and approximate
nearest-neighbour indexing \citep{johnson2019faiss} as the standard
implementation. Multi-hop benchmarks demand chained evidence: HotpotQA
\citep{yang2018hotpotqa} provides sentence-level supporting-fact
annotations across two-hop questions, and MuSiQue extends control over
reasoning structure through single-hop compositions
\citep{trivedi2022musique}. Surveys show that while RAG substantially
reduces hallucinations on knowledge-intensive tasks, retrieval errors
propagate through generation in ways aggregate accuracy scores cannot
readily surface \citep{gao2023ragsurvey,chen2023benchmarking}. Self-RAG
trains a single model to decide adaptively when to retrieve and to
critique its own evidence through self-reflection tokens
\citep{asai2024selfrag}, and dedicated multi-hop RAG evaluations show
that existing dense and sparse retrieval methods perform poorly when
several evidence hops are required \citep{tang2024multihoprag}. Beyond
single-model pipelines, structural analyses of multi-agent LLM
communication show that reasoning reliability depends on the topology
through which evidence propagates \citep{parks2026predictive},
reinforcing that what flows between components, not only what each
component emits, governs trust. ProvenAI does not train a new retriever
or generator; it measures what can be inferred around an inference-only
pipeline built from fixed pre-trained components
\citep{reimers2019sbert,qwen2025technical}.

\paragraph{Citation quality and verifiability.}
Generative search systems that produce fluent citation-bearing answers
can still fail to ground those citations in the corresponding source
passages \citep{liu2023verifiability}. FActScore decomposes long-form
generations into atomic claims and checks each against a knowledge
source \citep{min2023factscore}, RAGAS supplies reference-free metrics
for faithfulness and context relevance \citep{es2023ragas}, and the AIS
framework defines verifiable NLG as output attributable to identified
sources with a two-stage annotation pipeline \citep{rashkin2023measuring}.
ALCE formalised end-to-end evaluation of citation-bearing answers,
demonstrating that even strong models leave citations unsupported a
significant fraction of the time \citep{gao2023enabling}, and MIRAGE
uses model internals, specifically KL-divergence shifts, to attribute
answers to retrieved passages more faithfully than self-citation
prompting \citep{qi2024mirage}. ProvenAI is aligned with this
evaluation-focused direction but adds an orthogonal question: after a
document has been cited, does removing it from context actually change
what the model generates?

\paragraph{Context attribution, causal mediation, and provenance theory.}
Ablation-based attribution has a long lineage in machine-learning
interpretability, from local surrogate models such as LIME
\citep{ribeiro2016lime} to causal-mediation analyses that intervene on
internal model components \citep{vig2020mediation}, formalised through
the do-calculus \citep{pearl2009causality}. ContextCite formalises
context attribution by fitting a sparse linear surrogate over random
context ablations, with the key finding that removing high-scoring
context segments causes larger output probability drops than removing
low-scoring ones \citep{cohenwang2024contextcite}, and SelfCite
demonstrates that context ablation can serve directly as a
self-supervised reward for citation-generating models
\citep{chuang2025selfcite}. ProvenAI applies the same ablation intuition
at the document level and uses it to assign categorical influence labels
rather than to fine-tune a model. Provenance has also been advanced as a
structural security primitive in AI-driven systems, spanning secure
cloud architecture and migration \citep{alharthi2023scms} and
ontology-grounded forensic reasoning \citep{alharthi2025ciaf};
ProvenAI inherits this orientation and applies it to the citation layer
of retrieval-grounded QA. The broader notion of evidence provenance has
deep roots in the database literature, where formal frameworks
distinguish why, where, and how provenance
\citep{buneman2001why,cheney2009provenance}; we adopt this layered
perspective when discussing how ProvenAI's measurement infrastructure
composes with cryptographic enforcement (Section~\ref{sec:provnative}).
The Model Context Protocol standardises how AI applications access
external data and tools \citep{anthropic2024mcp,mcp2026intro}; ProvenAI
incorporates a local MCP layer as a traceable interface for
retrieval-time resource access without claiming that MCP itself resolves
attribution.

\section{Problem Formulation}
Let $q$ denote a question and $R = \{r_1, \dots, r_k\}$ the set of
retrieved resources provided to the model as context. The generator
produces a textual answer $a$ and a citation set $C \subseteq R$. The
dataset supplies a gold answer $a^{*}$ and a set of expected supporting
document titles $G$. ProvenAI evaluates three layers independently.
Answer correctness asks whether $a$ matches $a^{*}$ after lightweight
string normalisation. Citation fidelity asks whether the titles in $C$
align with the supporting titles in $G$. Resource influence asks whether
removing $r_i$ from context changes the answer or citation pattern.

\subsection{Citation fidelity}
Citation fidelity is operationalised at the document-title level using
token-set Jaccard similarity. Let $T_C$ denote cited titles and $T_G$
denote gold supporting titles. A cited title is counted as a match if
its Jaccard similarity to the nearest supporting title meets threshold
$\tau$:
\begin{align}
P_\tau &= \frac{1}{|T_C|}\sum_{t \in T_C}
  \indfn\!\left(\max_{g \in T_G}\mathrm{sim}(t,g)\ge \tau\right), \\
R_\tau &= \frac{1}{|T_G|}\sum_{g \in T_G}
  \indfn\!\left(\max_{t \in T_C}\mathrm{sim}(t,g)\ge \tau\right), \\
F_\tau &= \frac{2 P_\tau R_\tau}{P_\tau + R_\tau}.
\end{align}
The threshold $\tau$ is held fixed across the validation split
(Table~\ref{tab:config}); we report results under both exact
($\tau=1$) and semantic ($\tau=0.34$) matching.

\subsection{Resource influence: KL target and surface proxy}
\label{sec:influence}
The conceptual target for per-document influence is the KL divergence
between the full-context output distribution and the distribution
obtained after removing $r_i$:
\begin{equation}
\mathrm{Influence}_{\mathrm{KL}}(r_i) =
\KL\!\big(p(\cdot \mid q, R)\,\big\|\,p(\cdot \mid q, R \setminus \{r_i\})\big).
\label{eq:klinf}
\end{equation}
Equation~\ref{eq:klinf} is a do-calculus quantity in the sense of
\citet{pearl2009causality}: leave-one-resource-out ablation is the
intervention $\mathrm{do}(R \leftarrow R \setminus \{r_i\})$, and
$\mathrm{Influence}_{\mathrm{KL}}$ measures the natural direct effect of
$r_i$ on the output distribution \citep{vig2020mediation}. Because the
local MLX inference path used in this work does not expose per-token
probabilities, the system computes a surface-level proxy from
regenerated samples:
\begin{align}
\Delta a(r_i) &= 1 - \Jac\!\big(\mathrm{tok}(a),\mathrm{tok}(a_{-i})\big),\\
\Delta c(r_i) &= 1 - \frac{|C \cap C_{-i}|}{|C \cup C_{-i}|},\\
\phi(r_i) &= 0.8\,\Delta a(r_i) + 0.2\,\Delta c(r_i),
\end{align}
where $a_{-i}$ and $C_{-i}$ are the answer and citation set regenerated
after removing $r_i$. The proxy aggregates two surface signals:
token-level shift in the realised answer and Jaccard distance between
citation sets. The $0.8/0.2$ weighting reflects the relative reliability
of the two signals; principled probability-aware estimators
\citep{cohenwang2024contextcite,chuang2025selfcite} are deferred to a
backend that exposes per-token logits.

\begin{proposition}[Faithfulness of the surface proxy under
near-deterministic decoding]
\label{prop:faith}
Let $p_t$ and $p'_t$ denote the next-token distributions under the full
and ablated contexts at step $t$, conditioned on the same generated
prefix, and let $a, a_{-i}$ be the corresponding greedy decodings.
Suppose at every step $t$ before the first divergence the model is
$\epsilon$-confident, $\max_w p_t(w) \ge 1-\epsilon$ for some
$\epsilon \in (0,\tfrac12)$. If $\Delta a(r_i) > 0$, then at the first
step $t^{*}$ where $a_{t^{*}} \neq a_{-i,t^{*}}$,
\begin{equation}
\KL\!\big(p_{t^{*}}\,\big\|\,p'_{t^{*}}\big)\ \ge\
2\left(\tfrac12 - \epsilon\right)^{2}.
\label{eq:bound}
\end{equation}
In particular, in the deterministic-decoding limit $\epsilon \to 0$,
every observable answer change is witnessed by a strictly positive KL gap
of at least $\tfrac12$ nat, so $\Delta a$ is a sound (one-sided)
indicator of the KL target.
\end{proposition}

The reverse direction does not hold: KL shifts that preserve the argmax
leave $\Delta a$ at zero, the regime in which the proxy systematically
underestimates influence. A short proof appears in
Appendix~\ref{app:proof}. Proposition~\ref{prop:faith} clarifies the
empirical scope of the proxy: when decoding is near-deterministic,
$\Delta a > 0$ certifies a non-trivial KL shift, but the converse fails
because probability mass can redistribute over near-tied candidates
without altering the realised answer string. The citation component
$\Delta c$ partially compensates by detecting distributional shifts that
surface in the citation set even when the answer string is stable, but it
does not close the gap; closing it requires probability-aware backends.

Three practical consequences of the bound are worth stating explicitly.
First, every \textsc{used} or \textsc{hallucinated citation} verdict
supported by $\phi(r_i) > 0$ corresponds to a token-level KL shift of at
least $2(\tfrac12 - \epsilon)^2$ nats at some decoding step, so positive
proxy readings should be read as evidence of distributional displacement
rather than as artefacts of stochastic decoding. Second, the bound is
one-sided, so a \textsc{low influence} verdict cannot be read as a
certificate that $r_i$ had no effect on the output distribution; it
merely says the realised answer and citation set were stable under
removal. Third, because the bound depends only on argmax confidence
rather than on absolute log-probabilities, the proxy degrades gracefully
as the generator becomes less confident: the empirical $\Delta a$ will
increase under higher-temperature decoding even when
$\mathrm{Influence}_{\mathrm{KL}}$ is constant, which we control by
holding decoding deterministic across the validation split.

\subsection{Verdict assignment}
Each retrieved resource receives one of four labels via a measurable
function $V$ defined on the product of the citation indicator and the
influence proxy. With threshold $\theta$, $V$ assigns \textsc{used} when
$r_i \in C$ and $\phi(r_i) \ge \theta$; \textsc{hallucinated citation}
when $r_i \in C$ and $\phi(r_i) < \theta$; \textsc{uncited influential}
when $r_i \notin C$ and $\phi(r_i) \ge \theta$; and \textsc{low
influence} otherwise. The threshold is set adaptively per example as
$\theta = \max\!\big(\mathrm{median}(\{\phi(r_i)\}), \theta_{\mathrm{floor}}\big)$,
ensuring that approximately half of the documents are flagged as
influential while the floor protects against degenerate cases in which
all proxy values collapse to zero. The diagonal labels \textsc{used} and
\textsc{low influence} are the cases in which citation and influence
agree; the off-diagonal labels are the cases the rest of the pipeline is
designed to surface, and they are the operational instantiation of the
citation-influence gap.

\section{System Design}
\subsection{Seven-stage pipeline}
\begin{figure}[h]
    \centering
    \includegraphics[width=0.86\linewidth]{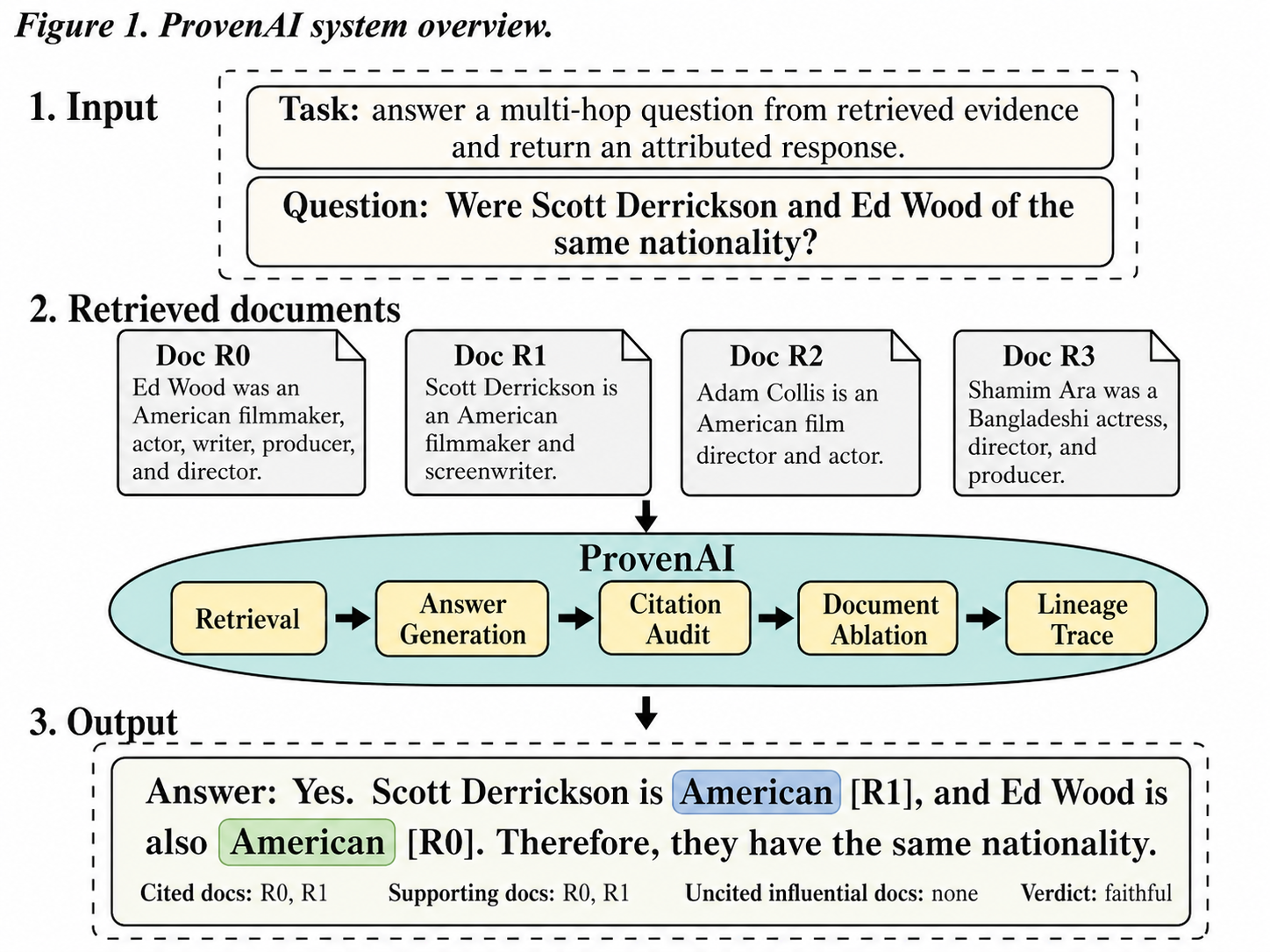}
    \caption{ProvenAI end-to-end system overview. A multi-hop question and
its retrieved evidence enter the pipeline; seven audit stages produce
structured outputs that include validation count, answer accuracy,
citation fidelity, corpus size, per-phase reports, and ablation verdict
tallies.}

\label{fig:overview}
\end{figure}

ProvenAI is organised into seven sequential stages summarised in
Table~\ref{tab:stages}, with the end-to-end flow visualised in
Figure~\ref{fig:overview}. The modular layout loosely couples data
preparation, retrieval, generation, attribution, ablation, evaluation,
and inspection so that individual stages can be tested, rerun, or swapped
independently, and each stage writes artefacts to disk that downstream
stages can read or reuse without recomputation.

\begin{table}[t]
\centering
\caption{ProvenAI pipeline stages and their saved outputs.}
\label{tab:stages}
\small
\begin{tabular}{@{}clp{0.45\linewidth}@{}}
\toprule
Stage & Primary role & Saved output \\
\midrule
1 & Normalise HotpotQA distractor records & JSONL examples, evidence files, manifest \\
2 & Build retrieval corpus and FAISS index & Canonical corpus, embeddings, FAISS index, SQLite lookup \\
3 & Generate citation-aware answers & Answer reports with retrieved resources \\
4 & Audit citation fidelity & Per-example citation metrics and verdicts \\
5 & Estimate document influence & Ablation reports and influence labels \\
6 & Run aggregate evaluation & Validation metrics and summary JSON \\
7 & Inspect results interactively & Streamlit views and exportable reports \\
\bottomrule
\end{tabular}
\end{table}

\subsection{Data, retrieval, and generation layers}
HotpotQA distractor records are converted into local JSONL files with
stable fields for example identifier, question text, gold answer,
supporting facts, supporting titles, and evidence documents. A
deterministic development subset (250 examples, seed 42) supports faster
iteration without rerunning the full validation split. The retrieval
corpus deduplicates evidence rows on title-text identity, removing
464{,}067 redundant copies while preserving provenance links and
yielding a canonical corpus of 509{,}300 passages. Document embeddings
are computed with the SBERT-style \texttt{all-MiniLM-L6-v2} model
\citep{reimers2019sbert} and indexed by FAISS with inner-product scoring
\citep{johnson2019faiss}; a SQLite mirror lets the inspection dashboard
serve records without loading the full JSONL file.

Each generation prompt elicits three sections (final answer, evidence
summary, chain of reasoning) with inline citation tags \texttt{[R0]},
\texttt{[R1]}, \ldots\ linking claims to retrieved resources. The default
local inference path uses \texttt{Qwen/Qwen2.5-3B-Instruct}
\citep{qwen2025technical}; a Hugging Face Transformers path is reserved
for future experiments requiring per-token probabilities. The output
parser repairs common formatting failures, extracts inline citation
references, retains raw model output alongside the parsed result, and
saves a structured JSON report per example. A local MCP server
\citep{anthropic2024mcp} exposes HotpotQA-backed tools through a standard
interface, recording JSON-RPC requests, returned resources, deterministic
resource identifiers, and trace graphs that can be replayed later. The
MCP layer is additive: it supplements attribution and ablation metrics
with an inspectable record of which resources entered the pipeline.

\section{Experimental Setup}
The evaluation covers the full HotpotQA distractor validation split.
Table~\ref{tab:config} reports dataset statistics and the primary runtime
configuration drawn directly from the saved artefacts. These values are
reported in full because modest changes to retrieval depth, model scale,
or citation threshold can shift both answer accuracy and audit behaviour
in non-trivial ways. Answer correctness is evaluated via lightweight
normalisation in which citation tags are stripped, strings are lowercased
and trimmed, and yes/no questions are handled with prefix matching; this
is adequate for internal diagnostics but should not be compared directly
against official HotpotQA leaderboard scores.

\begin{table}[t]
\centering
\caption{Dataset and corpus statistics (left) and primary runtime
configuration (right).}
\label{tab:config}
\small
\begin{minipage}[t]{0.48\linewidth}\centering
\begin{tabular}{@{}lr@{}}
\toprule
\multicolumn{2}{c}{Dataset and corpus}\\
\midrule
Train examples & 90{,}447 \\
Validation examples & 7{,}405 \\
Total examples & 97{,}852 \\
Train evidence rows & 899{,}667 \\
Validation evidence rows & 73{,}700 \\
Source evidence rows & 973{,}367 \\
Canonical retrieval rows & 509{,}300 \\
Duplicate rows removed & 464{,}067 \\
Development subset size & 250 \\
Random seed & 42 \\
\bottomrule
\end{tabular}
\end{minipage}\hfill
\begin{minipage}[t]{0.48\linewidth}\centering
\begin{tabular}{@{}lr@{}}
\toprule
\multicolumn{2}{c}{Runtime configuration}\\
\midrule
Benchmark & HotpotQA distractor \\
Retrieval depth & 10 \\
Embedding model & all-MiniLM-L6-v2 \\
Index & FAISS IP + SQLite \\
Generation backend & MLX (Apple Silicon) \\
Generation model & Qwen2.5-3B-Instruct \\
Citation limit & 2 \\
Prompt doc limit & 900 chars \\
Threshold $\tau$ & 0.34 \\
Influence weights & 0.8 / 0.2 \\
\bottomrule
\end{tabular}
\end{minipage}
\end{table}

\section{Results}
\subsection{Aggregate validation metrics}
Table~\ref{tab:agg} reports aggregate metrics over the full validation
split. Leave-one-resource-out ablation is disabled during the full-batch
run because, at retrieval depth 10, ablation per example requires
approximately $k+1 = 11$ forward passes per item, increasing total
inference cost by roughly an order of magnitude. The headline
observation is that citation fidelity exceeds answer accuracy by a
substantial margin (71.55\% versus 53.53\%): the model frequently
identifies and cites the correct supporting titles even on examples where
the final answer string is scored as incorrect. This divergence supports
the core argument of the paper. Answer correctness and evidence
alignment are not interchangeable, and collapsing them into a single
score would obscure cases where evidence handling is sound but answer
generation falls short, or vice versa. The zero aggregate influence score
reflects the decision to disable ablation during the full-validation run,
not a substantive empirical finding; influence estimates are examined
instead in targeted example-level reports.

\begin{table}[t]
\centering
\caption{Aggregate metrics over 7{,}405 HotpotQA distractor validation
examples.}
\label{tab:agg}
\small
\begin{tabular}{@{}lr@{}}
\toprule
Metric & Value \\
\midrule
Validation examples & 7{,}405 \\
Answer accuracy & 0.5353 \\
Mean citation-fidelity score & 0.7155 \\
Mean exact citation F1 & 0.7155 \\
Mean semantic citation F1 & 0.7155 \\
Aggregate influence score (ablation disabled) & 0.0000 \\
\bottomrule
\end{tabular}
\end{table}

The 18-point spread between citation fidelity and answer accuracy is the
population-level signature of the same decoupling that the case study
isolates at the example level. Roughly speaking, if citation fidelity and
answer accuracy were tightly coupled we would expect the ratio of
correct-and-faithful answers to total faithful answers to track the
marginal accuracy, and the two metrics would move together under any
change of generation backend or retrieval depth. Empirically they do
not. The model retrieves and cites the correct supporting titles
substantially more often than it produces a normalisation-matched answer
string, which means a non-trivial fraction of the validation set sits in
the regime where the evidence layer is sound but the answer-generation
layer fails downstream, plausibly through arithmetic errors on numeric
questions, format mismatches on yes/no items, or surface-level paraphrase
divergence. Conversely, a smaller but non-zero fraction sits in the
opposite regime, with a correct answer paired with a degraded citation
set, which is the population-level analogue of a \textsc{hallucinated
citation} verdict. The three-layer decomposition is what makes either
failure mode visible.

\subsection{Case study: the citation-influence gap}
\begin{figure}[h]
    \centering
    \includegraphics[width=0.86\linewidth]{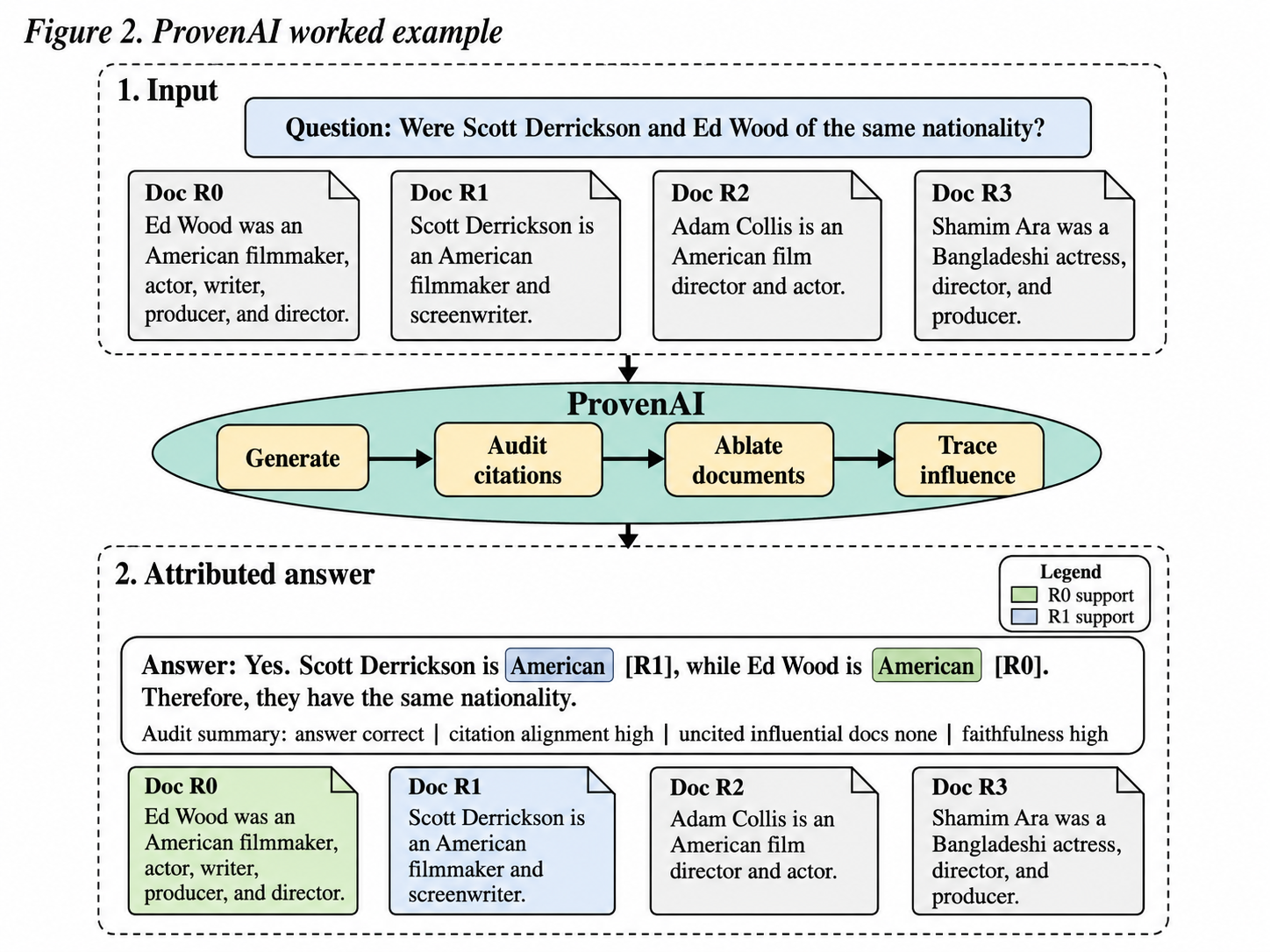}
    \caption{Worked example illustrating the citation-influence gap. The
generated answer cites the two titles identified by HotpotQA's
supporting-fact annotations and receives a clean citation audit, but
document-level ablation reveals that several uncited documents measurably
shift the output while one cited document registers only weak influence
under the current proxy.}

\label{fig:worked}
\end{figure}

A saved ablation report addresses the question \emph{Were Scott
Derrickson and Ed Wood of the same nationality?} The gold answer is yes;
the model correctly identifies both as American and cites the
corresponding retrieved documents. The citation audit is perfect: cited
titles match HotpotQA supporting titles exactly, yielding $F_\tau = 1.0$
under both exact and semantic matching. If the analysis stopped there,
the example would appear fully transparent. The influence report reveals
a more complicated picture. Ten retrieved documents are tested via
leave-one-out ablation. Figure~\ref{fig:worked} presents the example in
human-readable form, and Table~\ref{tab:verdicts} expresses the verdict
distribution numerically. One cited source falls below the influence
threshold while seven uncited sources each shift the answer or citation
set when removed. A plausible reading is that co-retrieved passages about
other American filmmakers stabilise the nationality pattern in the
model's output, even though none of those passages appear in the
response. This interpretation is consistent with prior observations that
language models can be perturbed by irrelevant context
\citep{shi2023distraction} and that retrieved evidence often acts at a
coarser granularity than self-citation suggests
\citep{qi2024mirage,cohenwang2024contextcite}. The example does not
establish token-level causal dependence, but it demonstrates that the
observable citation set and the empirical sensitivity profile can diverge
substantially, which is the defining feature of the citation-influence
gap.

\begin{table}[t]
\centering
\caption{Ablation verdict distribution for the Scott Derrickson / Ed Wood
example (mean influence proxy $= 0.97$).}
\label{tab:verdicts}
\small
\begin{tabular}{@{}lr@{}}
\toprule
Verdict & Count \\
\midrule
Used & 1 \\
Hallucinated citation & 1 \\
Uncited-influential & 7 \\
Low-influence & 1 \\
\bottomrule
\end{tabular}
\end{table}


\section{From Measurement to Provenance-Native Reasoning}
\label{sec:provnative}
ProvenAI is a measurement infrastructure: it observes a
retrieval-grounded pipeline from the outside and assigns each retrieved
document one of four diagnostic verdicts. A complementary line of work,
exemplified by emerging cryptographic-provenance architectures for
autonomous scientific workflows, treats provenance as an enforcement
property in which each module commits to its inputs and outputs at
execution time so that downstream consumers can verify, after the fact,
that the data flowing into a conclusion was authentic and unmanipulated.
The three layers ProvenAI distinguishes have natural counterparts in such
architectures. Answer correctness corresponds to checking a final claim
against a ground truth where one exists. Citation fidelity corresponds to
verifying that the resources nominally referenced by a claim were
retrieved through the trusted interface and, in cryptographic settings,
were signed by the upstream module that produced them. Resource influence
corresponds to confirming that the data which actually shaped the output
is the same data the system claims to have used.

The citation-influence gap suggests that the third layer is non-trivial
even when the first two are clean. A retrieval-grounded pipeline can be
cryptographically sound at every boundary, with every retrieved document
authentic and every citation referring to a genuinely retrieved passage,
and still produce answers shaped by passages outside the cited set. For
autonomous discovery settings in which AI systems orchestrate
experimental decisions, generate analysis code, or extract claims from
prior literature, this gap is an attack surface in its own right: an
adversary who cannot poison the data or forge a signature may still shape
conclusions by influencing what is retrieved, since uncited co-retrieved
passages can stabilise particular output patterns. Defences that validate
inputs before they reach the model, such as ontology-driven
prompt-injection mitigation \citep{alharthi2025promptshield,alharthi2026securebydesign}, and
zero-trust policies that treat every request as untrusted until verified
\citep{alharthi2024ztarl}, address the upstream half of this problem;
ProvenAI's influence layer addresses the downstream half by exposing
which retrieved evidence actually moved the output. Provenance-native
reasoning therefore requires an audit trail that exposes not only
\emph{which} documents were nominally used but \emph{which} documents the
realised output was actually sensitive to, a distinction that maps
cleanly onto the database literature's separation of why and how
provenance \citep{buneman2001why,cheney2009provenance}. ProvenAI's
contribution to this broader programme is to show that the latter
quantity is measurable and non-redundant with the former, and to provide
an end-to-end pipeline in which ablation-based influence sits alongside
cryptographically inspectable retrieval traces (via MCP) within a single
reproducible report. We view this composability as the practical content
of \emph{verifiable reasoning} in retrieval-grounded systems: a property
that combines structural commitment at module boundaries with behavioural
sensitivity analysis at the resource level.

Concretely, an auditable record for a retrieval-grounded inference in an
autonomous discovery setting would expose at least four classes of
artefact for each generated claim. The first is the retrieval manifest,
identifying every $r_i \in R$ together with the resource identifier under
which it was obtained from the trusted interface; in our implementation
this is precisely the role of the MCP trace graph (Appendix~\ref{app:mcp}).
The second is the citation manifest, identifying the subset
$C \subseteq R$ that the generator nominally invoked. The third is the
verdict assignment $V$, mapping each $r_i$ into one of the four
categorical labels by combining its citation indicator with its influence
signal. The fourth is the proxy-to-target relationship under which the
influence signal was computed, including the decoding-confidence regime
captured by Proposition~\ref{prop:faith}; without this fourth element the
influence layer is uninterpretable across pipelines. ProvenAI emits the
first three artefacts directly and characterises the fourth analytically,
so a downstream consumer reviewing a synthetic scientific claim can audit
not only which sources were retrieved and cited but the conditions under
which the recorded influence numbers should be trusted. The same
auditability requirement has motivated ontology-grounded and
reinforcement-learning-based pipelines for cloud forensics and incident
response \citep{alharthi2025ciaf,alharthi2024incident,alharthi2025llmforensics},
where the value of an automated conclusion depends on a defensible record
of the evidence behind it.

\section{Discussion and Limitations}
\paragraph{Citations indicate claimed use, not confirmed influence.}
A citation tells the reader which source the answer nominally refers to
but does not confirm that this source was the primary driver of the
generated content. Self-citing systems frequently struggle to match
required citation formats, refer to non-existent sources, or fail to
reflect how context was actually used during generation
\citep{qi2024mirage,gao2023enabling}. ProvenAI treats citation as one
measurable layer and tests it against a second, independent layer: output
sensitivity under resource removal. The three-layer decomposition exposes
failure modes such as correct answer with poor citation fidelity, or
clean citation audit with diffuse influence, that aggregate accuracy
cannot distinguish.

\paragraph{Toward distributional influence measurement.}
The most significant limitation of the current implementation is that the
local MLX backend does not provide per-token probability outputs,
preventing direct computation of $\mathrm{Influence}_{\mathrm{KL}}$ from
Equation~\ref{eq:klinf}. Proposition~\ref{prop:faith} bounds the regime
in which the surface proxy is faithful but also identifies its blind
spot: when probability mass redistributes among ties, $\Delta a$ remains
zero. ContextCite's sparse linear surrogate over random ablations
\citep{cohenwang2024contextcite} and SelfCite's necessity-and-sufficiency
ablation rewards \citep{chuang2025selfcite} both rely on log-probability
access; replacing the local backend with an inference path that exposes
per-token probabilities would let ProvenAI adopt these probability-aware
estimators, moving from the current surface proxy toward distributional
measurement.

\paragraph{Scale, attribution granularity, and scope.}
The current local path uses a 3B Qwen2.5 model for stability, with
planned experiments targeted at 7B to 8B models on a higher-resource
environment. The smaller model likely yields lower answer accuracy and
may affect both citation behaviour and influence estimates in ways that
would not generalise to larger systems. The attribution audit also
operates at the document-title level: it does not verify sentence-level
entailment or confirm that the specific passage within a cited document
actually supports the corresponding claim, so a title match is necessary
but not sufficient for faithful citation
\citep{rashkin2023measuring,min2023factscore}. All reported results
derive from the HotpotQA distractor split; before drawing broad
conclusions about the citation-influence gap, the phenomenon should be
examined on further benchmarks and open-domain settings where the
document pool is not curated per question.

\paragraph{Broader impact.}
ProvenAI is designed to improve calibration of trust in
retrieval-grounded systems. A plausible risk is that detailed audit
artefacts may paradoxically increase over-reliance: users who see rich
trace outputs may treat them as proof of correctness, a calibration risk
familiar from work on security awareness and human-facing trust
mechanisms \citep{alharthi2021social}. The interface should therefore
present influence scores explicitly as diagnostic indicators, not
guarantees. The work uses a publicly available benchmark and involves no
collection of human-subject data.

\section{Conclusion}
ProvenAI demonstrates that transparency in retrieval-grounded QA is
better understood as a layered property than as a single indicator. The
generated answer, the citation list, and the resource influence profile
each reveal complementary information, and they can disagree in ways a
single correctness score would hide. On the HotpotQA distractor
validation split, the system achieves moderate answer accuracy alongside
substantially higher citation fidelity, and the ablation case study shows
that a clean citation audit can coexist with a complex, partially uncited
influence profile, the pattern we call the citation-influence gap. We
have formalised the relationship between the implemented surface proxy
and a token-level KL-divergence target, situated the framework within
causal-mediation analysis and database-provenance theory, and argued that
measurement-based audit composes naturally with the
cryptographic-provenance architectures that autonomous discovery settings
increasingly require. Limitations include a proxy-based influence
estimate, a model smaller than the original design target, and
title-level attribution as only a first step toward full evidence
verification. ProvenAI nonetheless provides a reproducible foundation for
auditable multi-hop QA research and a clear path toward more rigorous,
probability-aware provenance analysis.

\bibliography{references-26}

\appendix
\section{Proof of Proposition~\ref{prop:faith}}
\label{app:proof}
We restate the setting briefly. Let $p_t, p'_t$ be the next-token
distributions under the full and ablated contexts at step $t$,
conditioned on the same generated prefix, and let $a, a_{-i}$ be the
corresponding greedy decodings. Assume
$\max_w p_t(w) \ge 1-\epsilon$ for all $t$ before the first divergence,
with $\epsilon \in (0,\tfrac12)$. Suppose $\Delta a(r_i) > 0$, so
$a \neq a_{-i}$, and let $t^{*}$ be the first step where
$a_{t^{*}} \neq a_{-i,t^{*}}$. Write
$w := a_{t^{*}} = \argmax_u p_{t^{*}}(u)$ and
$w' := a_{-i,t^{*}} = \argmax_u p'_{t^{*}}(u)$ with $w \neq w'$. By the
confidence assumption, $p_{t^{*}}(w) \ge 1-\epsilon$. Because $w'$ is the
argmax under $p'_{t^{*}}$ and $w \neq w'$, it must hold that
$p'_{t^{*}}(w) \le p'_{t^{*}}(w')$, and since these are two disjoint
masses summing to at most one, $p'_{t^{*}}(w) \le \tfrac12$.

The total variation distance is bounded below by the gap at coordinate
$w$:
\begin{equation}
\|p_{t^{*}} - p'_{t^{*}}\|_{\mathrm{TV}} \ \ge\
p_{t^{*}}(w) - p'_{t^{*}}(w) \ \ge\ (1-\epsilon) - \tfrac12
= \tfrac12 - \epsilon.
\end{equation}
Pinsker's inequality,
$\KL(p \,\|\, q) \ge 2\|p - q\|_{\mathrm{TV}}^{2}$, then yields
\begin{equation}
\KL\!\big(p_{t^{*}}\,\big\|\,p'_{t^{*}}\big)\ \ge\
2\left(\tfrac12 - \epsilon\right)^{2}.
\end{equation}
As $\epsilon \to 0$ the bound approaches $\tfrac12$ nat, establishing the
claimed one-sided witnessing.

For the converse, consider $p_{t^{*}}(w) = 0.9$, $p_{t^{*}}(u) = 0.1$ and
$p'_{t^{*}}(w) = 0.6$, $p'_{t^{*}}(u) = 0.4$ for some $u \neq w$. Both
distributions have argmax $w$, hence $a_{t^{*}} = a_{-i,t^{*}}$ and
$\Delta a$ contributes zero at this step, yet
$\KL(p_{t^{*}} \,\|\, p'_{t^{*}}) \approx 0.092$ nat is strictly
positive. The proxy is therefore one-sided: positive answer-change
implies positive KL, but not conversely.

\section{Reproducibility Details}
\label{app:repro}
The repository organises ProvenAI into scriptable phases. The
representative commands below reproduce the main artefact pipeline once
the HotpotQA cache and required dependencies are available:
\begin{quote}\ttfamily\small
PYTHONPATH=src python3 scripts/prepare\_hotpotqa.py\\
PYTHONPATH=src python3 scripts/build\_faiss\_index.py\\
PYTHONPATH=src python3 scripts/run\_experiment.py\\
PYTHONPATH=src python3 scripts/run\_attribution\_audit.py\\
PYTHONPATH=src python3 scripts/run\_ablation.py\\
PYTHONPATH=src python3 scripts/run\_evaluation.py\\
PYTHONPATH=src python3 scripts/run\_mcp\_server.py
\end{quote}
The current configuration records the dataset variant, retrieval depth,
generation backend, attribution threshold, evaluation split, and report
locations. The latest saved artefacts referenced in this draft are the
manifest under \path{data/processed/manifest.json}, the retrieval
metadata under
\path{data/index/hotpotqa_distractor_hashing_metadata.json}, the
attribution audit under \path{reports/phase4/latest_audit.json}, the
ablation report under \path{reports/phase5/latest_ablation.json}, and
the aggregate evaluation under
\path{reports/phase6/latest_evaluation.json}.

\section{Portal and Dashboard Screenshots}
\label{app:portal}
The interactive ProvenAI portal supports inspection of generated answers,
retrieved evidence, citation fidelity, ablation behaviour, and MCP trace
lineage. The screenshots below render once the corresponding image files
are uploaded.

\begin{figure}[h]
\centering

  {\includegraphics[width=\linewidth]{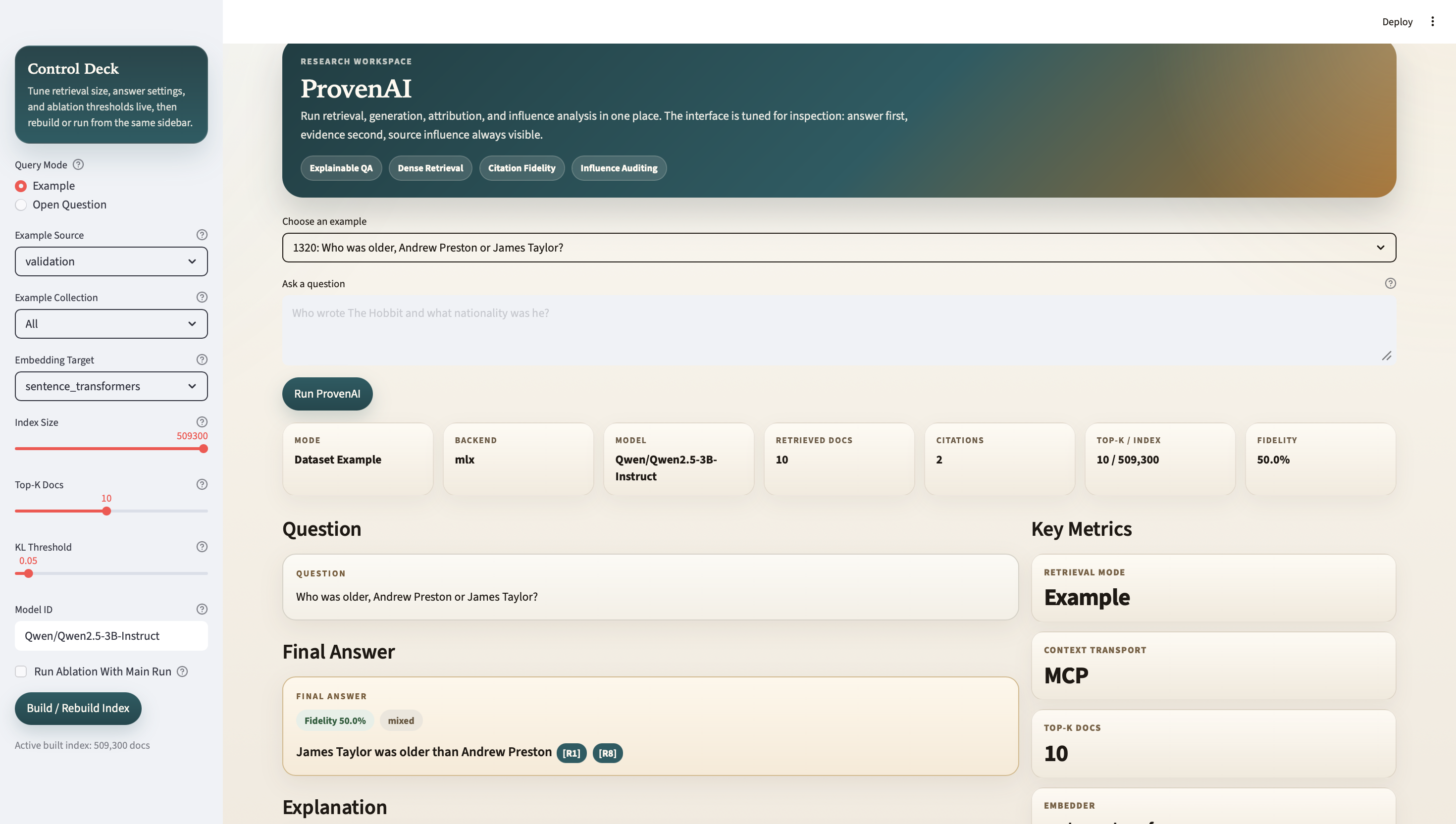}}
 
\caption{ProvenAI console overview showing the question, generated
answer, citation count, and evaluation summary.}
\label{fig:console}
\end{figure}

\begin{figure}[h]
\centering
  {\includegraphics[width=\linewidth]{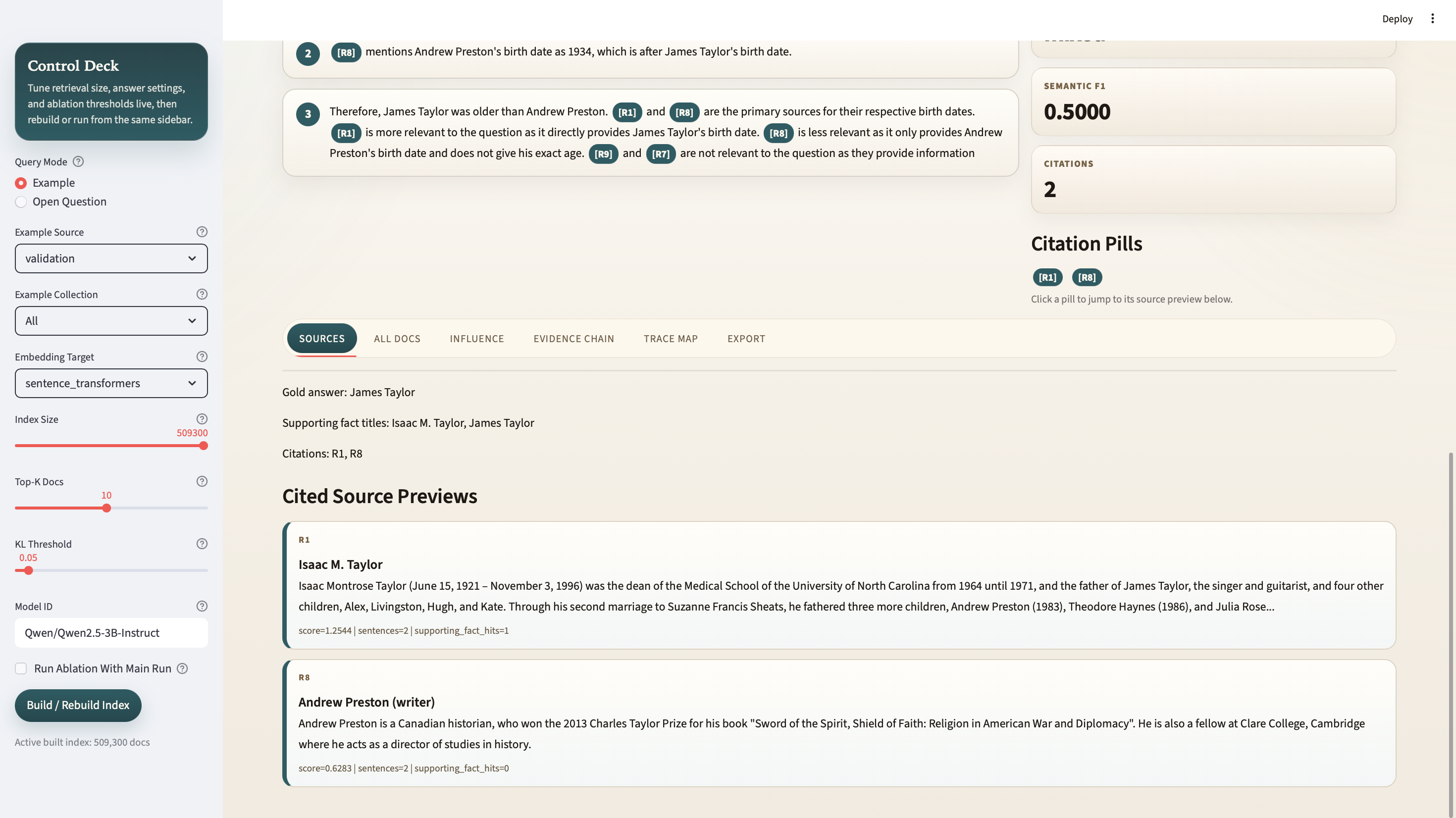}}
 
\caption{Evidence inspection view showing retrieved documents, cited
sources, and supporting context.}
\label{fig:evidence}
\end{figure}

\section{Additional Experimental Details}
\label{app:addl}
ProvenAI is configured for the HotpotQA distractor setting. The current
draft reports the saved full-validation evaluation with 7{,}405 examples,
top-$k = 10$ retrieval, semantic attribution threshold 0.34, and the
local Apple Silicon MLX generation path. The repository also supports
smaller deterministic development runs for faster iteration. The Phase 5
influence analysis should be interpreted as a proxy measurement in the
current implementation. The local MLX path records answer and citation
changes under leave-one-resource-out ablation but does not expose
token-level probabilities for KL-divergence influence scoring; claims
about resource influence in this draft should therefore be framed as
answer-and-citation sensitivity rather than probability-level causal
influence, as quantified by Proposition~\ref{prop:faith}.

\section{MCP Traceability}
\label{app:mcp}
The MCP integration exposes retrieval and evidence resources through
structured tool calls and stable resource identifiers. Representative MCP
operations include \texttt{get\_example},
\texttt{get\_supporting\_facts}, \texttt{search\_hotpotqa},
\texttt{get\_evidence\_document}, \texttt{get\_trace\_session\_status},
and \texttt{get\_trace\_graph}. These traces link retrieval calls,
returned resources, cited evidence, and downstream reports, making
provenance inspectable after generation.

\begin{figure}[h]
\centering
  \includegraphics[width=\linewidth]{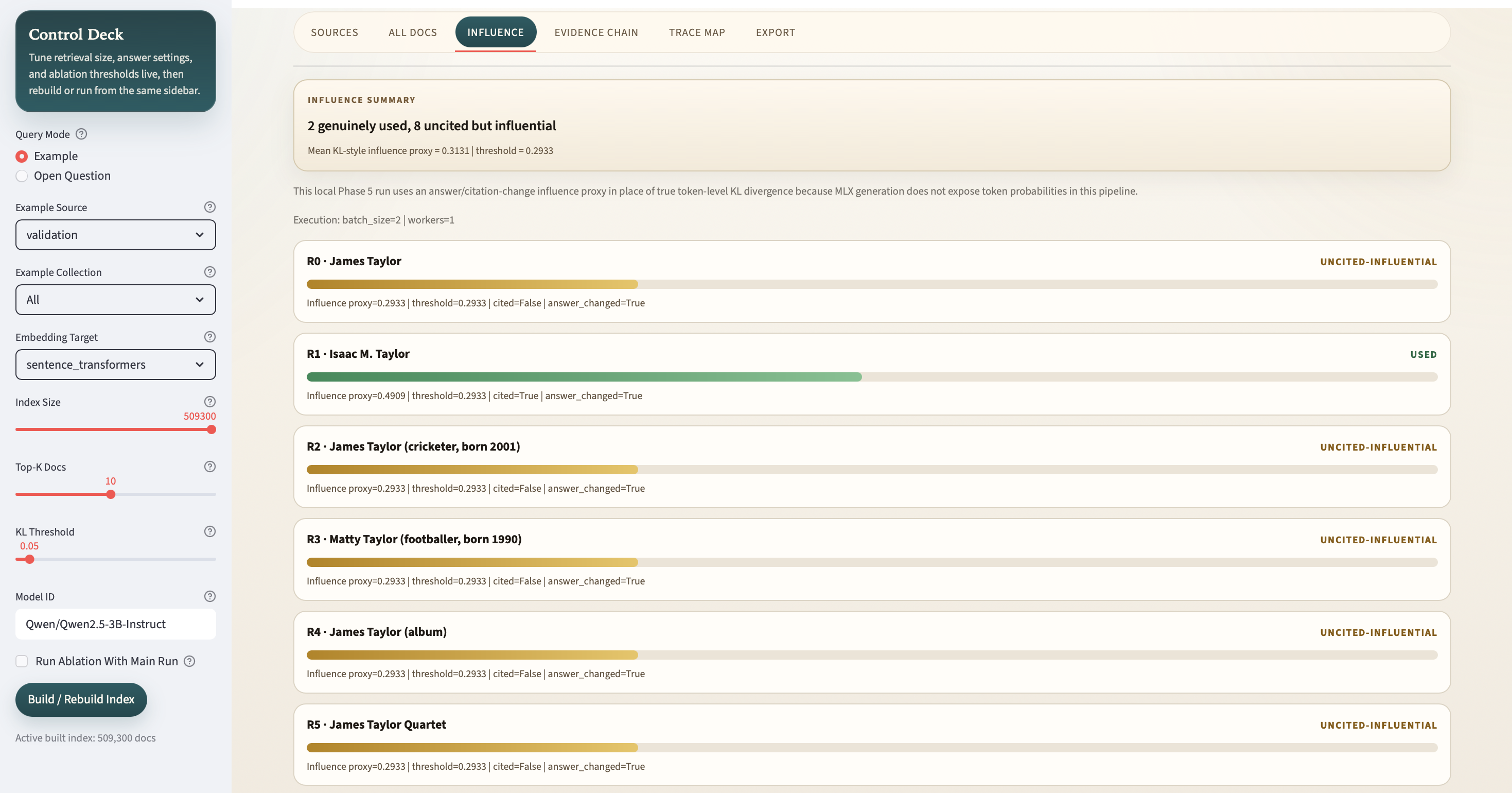}

\caption{Ablation view showing leave-one-resource-out influence results and answer or citation changes.}
\label{fig:ablation}
\end{figure}

\end{document}